\documentclass[10pt, journal, cspaper, compsoc]{IEEEtran}
\usepackage[breaklinks=true, colorlinks, citecolor=blue, linkcolor=blue, urlcolor=blue]{hyperref}
\usepackage{url}            
\usepackage{booktabs}       
\usepackage{amsfonts}       
\usepackage{nicefrac}       
\usepackage{subfigure}
\usepackage{times}
\usepackage{graphicx}
\usepackage{amsmath}
\usepackage{amssymb}
\usepackage{color}
\usepackage{xcolor}
\usepackage{multirow}
\usepackage{overpic}
\usepackage{amssymb}
\usepackage{pifont}
\usepackage{makecell}
\usepackage{tablefootnote}
\usepackage[nocompress]{cite} 
\usepackage{ragged2e}
\usepackage{cancel}
\usepackage{marvosym}

\usepackage{geometry}
\def\eg{\emph{e.g., ~}}
\def\ie{\emph{i.e., ~}}
\def\etal{{\em et al.}}

\renewcommand{\eqref}[1]{Equ.~(\ref{#1})}
\newcommand{\thRows}[1]{\multirow{3}*{#1}}

\newcommand{\tRows}[1]{\multirow{2}*{#1}}
\newcommand{\tCols}[1]{\multicolumn{2}{c}{#1}}

\newcommand{\myPara}[1]{\vspace{.15in} \noindent\textbf{#1}}
\newcommand{\figref}[1]{Fig.~\ref{#1}}
\newcommand{\tabref}[1]{Tab.~\ref{#1}}
\newcommand{\secref}[1]{Section~\ref{#1}}
\usepackage{bm}

\newcommand{\red}[1]{{\textcolor{red}{#1}}}
\newcommand{\bwy}[1]{{{#1}}}
\newcommand{\reversion}[1]{{{#1}}}
\newcommand{\ourMthd}{HSSL}
\newcommand{\tb}[1]{\textbf{#1}}
\usepackage{subfigure}

\usepackage{silence}
\graphicspath{{./figures/}}

\newcommand{\tablestyle}[2]{\setlength{\tabcolsep}{#1}\renewcommand{\arraystretch}{#2}\centering}

\def\ViT{ViT \cite{dosovitskiy2020vit}}
\def\Swin{Swin \cite{liu2021Swin}}
\def\PoolFormer{PoolFormer \cite{yu2022metaformer}}
\def\ResNet{ResNet \cite{he2016deep}}
\def\ResMLP{ResMLP \cite{touvron2021resmlp}}
\def\ConvNext{ConvNext \cite{liu2022convnet}}
\def\MoCo{MoCo \cite{Chen_2021_ICCV}} 
\def\DINO{DINO \cite{caron2021emerging}} 
\def\iBOT{iBOT \cite{zhou2021ibot}}
\def\MAE{MAE \cite{he2021masked}}
\def\MFF{MFF \cite{Liu_2023_ICCV}}
\def\AttMask{AttMask \cite{attmask}}

\begin{document}

\title{Enhancing Representations through Heterogeneous Self-Supervised Learning}

\author{Zhong-Yu Li, Bo-Wen Yin, Yongxiang Liu, Li Liu, Ming-Ming Cheng\textsuperscript{\Letter}
\IEEEcompsocitemizethanks{\IEEEcompsocthanksitem 
Z.-Y. Li, B.-W. Yin, and M.-M. Cheng are with VCIP \& TBI Center, 
Nankai University, Tianjin 300350, China. 
\IEEEcompsocthanksitem 
M.-M. Cheng is also with NKIARI, Futian, Shenzhen, China. 
\IEEEcompsocthanksitem Y Liu and L Liu are with College of Electronic Science 
and Technology, National University of Defense Technology (NUDT), Changsha, China. 
\IEEEcompsocthanksitem \textsuperscript{\Letter}Corresponding author: Ming-Ming Cheng.
\IEEEcompsocthanksitem This work was partially supported by the National Key 
Research and Development Program of China No. 2021YFB3100800, 
and the National Natural Science Foundation of China (62225604, 62376283).
Computation is supported by the Supercomputing Center of Nankai University.
}}


\IEEEtitleabstractindextext{
\begin{abstract} \justifying
Incorporating heterogeneous representations from different architectures has facilitated various vision tasks, \eg some hybrid networks combine transformers and convolutions. 
However, complementarity between such heterogeneous architectures has not been well exploited in self-supervised learning. 
%
Thus, we propose \tb{H}eterogeneous \tb{S}elf-\tb{S}upervised \tb{L}earning~(\tb{\ourMthd{}}), 
which enforces a base model to learn from an auxiliary head whose architecture is heterogeneous from the base model. 
In this process, \ourMthd~endows the base model with new characteristics in a representation learning way without structural changes.
To comprehensively understand the \ourMthd{}, we conduct experiments on various heterogeneous pairs containing a base model and an auxiliary head. 
We discover that the representation quality of the base model moves up as their architecture discrepancy grows. 
This observation motivates us to propose a search strategy that quickly determines the most suitable auxiliary head for a specific base model to learn and several simple but effective methods to enlarge the model discrepancy. 
The \ourMthd{} is compatible with various self-supervised methods, achieving superior performances on various downstream tasks, 
including image classification, semantic segmentation, 
instance segmentation, and object detection. 
The code and dataset are available at \url{https://github.com/NK-JittorCV/Self-Supervised/}.
\end{abstract}
}

\maketitle
\IEEEdisplaynontitleabstractindextext
\IEEEpeerreviewmaketitle

\IEEEraisesectionheading{\section{Introduction}\label{sec:introduction}}

\IEEEPARstart{S}{elf-supervised} learning has succeeded in learning rich representations without requiring expensive annotations. 
This success is attributed to different pretext tasks, especially instance discrimination \cite{caron2018deep, caron2021emerging, tian2020contrastive, PCL} and masked image modeling \cite{he2021masked, huang2024contrastive, baevski2022data2vec}.
Adapting these methods to various network architectures, \eg convolution neural network \cite{sere, caron2020unsupervised}, vision transformer \cite{caron2021emerging, sere, zhou2021ibot, byol} and Swin transformer \cite{liu2021Swin}, has brought superior performances on a variety of downstream tasks, including image classification \cite{russakovsky2015imagenet}, semantic segmentation \cite{Everingham2009ThePV, Xie_2021_ICCV} and object detection \cite{lin2015microsoft}.

Different neural network architectures learn representations with distinct characteristics that reveal the intrinsic properties of an architecture, \eg the global and local modeling abilities.
Prior works \cite{Zhu_2019_ICCV, yan2021contnet, ge2021revitalizing, yin2023dformer} have demonstrated that the characteristics of different architectures can be complementary. 
Section 1 of the supplementary material also provides a pilot experiment to demonstrate the superiority of combining different architectures over a single architecture.
Existing methods \cite{convit, gulati2020conformer, liu2021Swin, Pan_2022_CVPR} mainly focus on architecture design to leverage such complementarity. 
However, we utilize the complementarity in a representation learning way while not modifying the model architecture.

\begin{figure}[t]
  \centering
  \begin{overpic}[width=.8\linewidth]{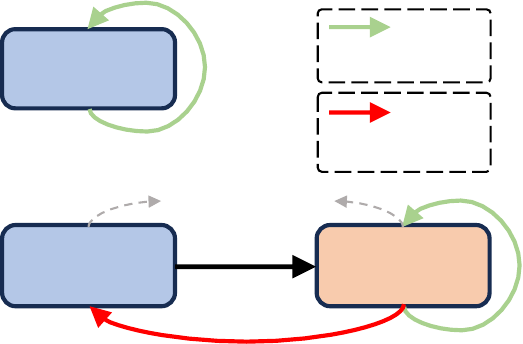}
    \put(6.0, 51.5){base model}
    \put(6.0, 13.5){base model}
    \put(63.5, 13.5){auxiliary head}
    \put(-2.5, 36.0){(a) existing methods}
    \put(33.0, -5.0){(b) \ourMthd~(Ours)}
    \put(62.0, 53.0){self-supervision}
    \put(62.0, 39.0){\tb{heterogeneous}}
    \put(62.0, 35.0){self-supervision}
    \put(32.5, 26.0){\tb{heterogeneous}}
  \end{overpic}
  \caption{Illustration of the heterogeneous 
    self-supervised learning~(\ourMthd).
    (a) General self-supervised learning methods make a base model supervise itself. 
    (b) The \ourMthd~supervises the base model under the 
    guidance of an auxiliary head 
    whose architecture is heterogeneous to the base model, 
    making the base model learn new characteristics.
  }\label{fig:introduction}
\end{figure}

Inspired by the above analysis, we propose {H}eterogeneous {Self}-Supervised {L}earning~(\ourMthd{}), which enhances a model with the characteristics of any other architectures. 
Specifically, during pre-training, the model comprises a base model and an auxiliary head whose architecture is heterogeneous to the base model.
Such heterogeneity makes the auxiliary head provide missing characteristics from the base model.
To endow the base model with its missing characteristics, we encourage the representations of the base model to mimic the representations of the auxiliary head, as shown in \figref{fig:introduction}.
Once pre-training is complete, the base model integrates new characteristics and we remove the auxiliary head.

For a comprehensive analysis, we examine various heterogeneous pairs of the base model and the auxiliary head and discover that the improvement in the base model is positively related to the discrepancy between the base model and the auxiliary head. 
A more significant discrepancy implies that the auxiliary head can provide more characteristics missing from the base model, thus magnifying the gains of the base model. 
This observation allows a specific base model to choose the most suitable auxiliary head. 
We propose a quick search strategy that simultaneously examines all candidate auxiliary heads to perform heterogeneous representation learning with the same base model. 
Thus, we can quickly determine the most suitable auxiliary head. 
Moreover, we further modify the chosen auxiliary head to enlarge its discrepancy with the base model to boost the performance.

Our proposed \ourMthd{} can be implemented in different self-supervised learning schemes, \eg contrastive learning \cite{Chen_2021_ICCV}, self-clustering \cite{caron2021emerging}, and masked image modeling \cite{he2021masked}, thus orthogonal to multiple self-supervised training methods \cite{Chen_2021_ICCV, caron2021emerging, zhou2021ibot, he2021masked}. 
On various downstream tasks, including image classification \cite{russakovsky2015imagenet}, semantic segmentation \cite{Everingham2009ThePV}, semi-supervised semantic segmentation \cite{gao2022luss, sun2023corrmatch}, instance segmentation \cite{lin2015microsoft}, and object detection \cite{lin2015microsoft, Everingham2009ThePV}, \ourMthd~consistently brings significant improvements for various network architectures without structure change. 

Our major contributions are summarized as follows:
\begin{itemize}
  \item We propose heterogeneous self-supervised learning, enabling a base model to learn the characteristics of different architectures.
  \item Through extensive experiments, we discovered that the discrepancy between the base model and the auxiliary head is positively related to the improvements in the base model and propose a quick search strategy to find the most suitable auxiliary head for a specific base model. 
  \item The proposed representation learning manner is compatible with existing self-supervised methods and consistently boosts performances across various downstream tasks. 
\end{itemize}

\section{Related Work}

\subsection{Self-Supervised Learning}

Self-supervised learning enables learning rich representations in the unsupervised setting, reducing the cost of collecting annotations. 
Early methods design different pretext tasks that can generate free supervision, such as coloration \cite{zhang2016colorful, Larsson_2017_CVPR}, jigsaw puzzles \cite{norooziECCV16}, rotation prediction \cite{gidaris2018unsupervised}, autoencoder \cite{Doersch_2015_ICCV, VincentLBM08}, image inpainting \cite{Pathak_2016_CVPR} and counting \cite{Noroozi_2017_ICCV}. 
The recent success of self-supervised learning can be attributed to instance discrimination \cite{Wu_2018_CVPR, Zhao_2021_ICCV, Dwibedi_2021_ICCV, Koohpayegani_2021_ICCV, dosovitskiy2014discriminative} and masked image modeling \cite{bao2021beit, he2021masked, Xie_2022_CVPR, Liu_2023_CVPR, chen2022efficient} methods. 
\reversion{New paradigms, such as correlational image modeling \cite{li2023correlational} and corrupted image modeling \cite{fang2023corrupted}, have been proposed, further enriching the field.
}

\myPara{Instance discrimination.} 
\reversion{Instance discrimination generates multiple views of an image through random image augmentations and aligns their representations \cite{roh2021scrl, Henaff_2021_ICCV, mugs2022SSL, dec2016, misra2020self}. 
This framework has been extended with various loss formulations, including contrastive learning \cite{DecoupledContrastive, chen2020simple, wang2020DenseCL, Xie_2021_CVPR, Xie_2021_ICCV}, feature alignment \cite{DynamicsContrastive, Chen_2021_CVPR, ermolov2021whitening}, clustering assignment \cite{Zhan_2020_CVPR, yangCVPR2016joint, oquab2023dinov2, Zhu_2023_ICCV, caron2019unsupervised}, redundancy reduction \cite{barlow_Twins, bardes2022vicreg}, sorting \cite{Shvetsova_2023_ICCV}, and relational modeling \cite{sere, gao2022towards}.  
These methods have been applied at both image-level \cite{chen2020simple, asano2020self, ijcai2022p200, Song_2023_ICCV} and dense-level \cite{zhang2022densesiam, o2020unsupervised, xie2021unsupervised, huang2022learning}, and demonstrate broad adaptability across architectures, including convolutional neural networks \cite{caron2020unsupervised}, vision transformers \cite{caron2021emerging}, and Swin Transformers \cite{liu2021Swin}.
However, existing approaches often overlook the potential complementarity between different architectures.
In this work, we propose \ourMthd, a framework designed to harness complementary characteristics across different architectures using a heterogeneous self-supervised learning scheme. 
Moreover, our method is orthogonal to existing self-supervised techniques.}

\myPara{Masked image modeling.}
\reversion{The masked image modeling~(MIM) based methods \cite{attmask, li2021mst, li2022efficient, hdm2023} reconstruct masked image patches based on the unmasked ones, emphasizing spatial context learning.
Researchers have explored diverse reconstruction targets to capture representations with varying properties.
For example, pixel-based reconstruction \cite{Xie_2022_CVPR, he2021masked, Feng_2023_CVPR, li2022semmae, xie2023masked, Liu_2023_ICCV} often yields strong yet non-linear representations. 
To endow representations with strong semantic information, more types of targets, \eg hand-designed HOG \cite{hog2005, Wei_2022_CVPR, Wang_2023_CVPR}, frequency \cite{xie2023masked, liu2023devil}, masked positions \cite{wang2023droppos}, features from online network \cite{gao2022towards, yi2022masked, tao2023siamese, Assran_2023_CVPR, chen2022sdae}, discretized tokens \cite{bao2021beit, chen2024context, dong2023peco}, or the combination of multiple targets \cite{gao2024mimic, chen2024context, dong2022bootstrapped}.
Recent research \cite{gao2022towards} also reconstructs representations from an off-the-shelf pre-trained model and achieves excellent performance, especially when using large-scale datasets \cite{Fang_2023_CVPR}. 
When using the online network \cite{liu2024exploring, chen2022sdae}, some works \cite{zhou2021ibot, huang2024contrastive, assran2022masked, wu2022extreme, jiang2023layer, shi2022adversarial} further combine the advantages of masked image modeling and instance discrimination to boost the performance. 
Meanwhile, apart from targets, some works \cite{zhang2023contextual, li2022semmae, hdm2023, attmask} also investigate different masking strategies to facilitate high-level representations.}

\reversion{Similar to instance discrimination, masked image modeling \cite{Woo2023ConvNeXtV2, huang2022green, zhou2021ibot, gao2022mcmae, Li2022ummae, tian2023designing, tian2023integrally, peng2022unified} has also been applied to diverse architectures like vision transformer \cite{dosovitskiy2020vit}, ConvNext-V2 \cite{Woo2023ConvNeXtV2}, and \Swin. 
These developments underscore the potential of leveraging architectural diversity to improve MIM-based representation learning.}

\begin{figure*}[t]
  \centering
  \begin{overpic}[width=.8\linewidth]{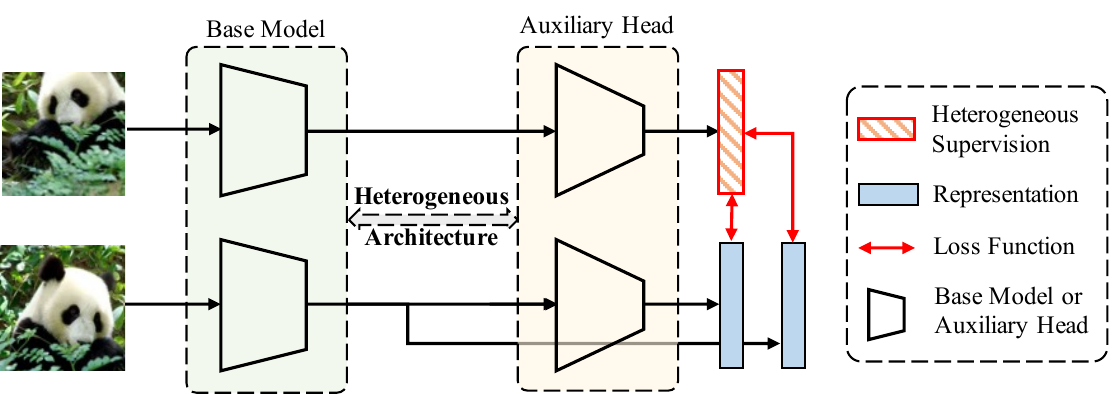}
    \put(22.4, 22.8){$f_1$}
    \put(4.5, 15.5){$x_1$}
    \put(4.5, 0.1){$x_2$}
    \put(22.4, 7){$f_2$}
    \put(52.7, 22.8){$h_1$}
    \put(52.7, 7){$h_2$}
    \put(64.9, 0){$z_2^h$}
    \put(64.9, 30.5){$z_1^h$}
    \put(70.5, 0){$z_2^{b}$}
    \put(32.5, 25){$z_1^{b}$}
    \put(32.5, 5.3){$z_2^{b}$}
  \end{overpic}
  \caption{Our \ourMthd~framework. 
    The architectures of the base model and the auxiliary head are heterogeneous. 
    The representations extracted by the auxiliary head supervise the two networks simultaneously. 
    The base model and the auxiliary head can be arbitrary architectures, such as \ViT, \Swin, \ConvNext, \ResNet, \ResMLP, and \PoolFormer.
  }\label{fig:vis_framework}
\end{figure*}

\subsection{Heterogeneity on Neural Network}

The heterogeneous neural network, which combines multiple types of architectures \cite{Yang_2022_CVPR, Pan_2022_CVPR, yin2023dformer}, can generate complementary characteristics and facilitate various vision tasks, including semantic segmentation \cite{Zhu_2019_ICCV, Fu_2019_CVPR, YUAN2023109228}, object detection \cite{yin2022camoformer, Li2022ExploringPV}, image classification \cite{mehta2022mobilevit, yan2021contnet, Wu_2021_ICCV}, and image quality assessment \cite{Lao_2022_CVPR}. 
These methods mainly design new architectures to leverage complementarity. 
For example, Wu~\etal \cite{Wu_2021_ICCV} combines convolution and attention in an architecture to achieve better classification accuracy. 
In comparison, we enforce a network constructed by a specific architecture to learn characteristics from any other architectures via representation learning without any structural changes. 
Thus, the proposed method is flexible in fusing characteristics from any architectures. 

Some works \cite{song2023multimode, ge2021revitalizing} have tried to utilize the complementarity 
to improve self-supervised learning. 
Specifically, these methods make the ViT and ResNet guide each other. 
However, beyond this pair, they lack a comprehensive analysis and understanding of the complementarity between different architectures. 
In comparison, we investigate a wide range of architectures, not only ViT and ResNet, and provide a comprehensive analysis of why and how the complementarity benefits self-supervised learning. 
We discover that a more significant model discrepancy leads to more significant improvements, enabling us to design more suitable auxiliary heads to guide a specific model.

\section{Method}
\label{sec:method}

In \secref{sec:Preliminaries}, we recall the existing self-supervised methods. 
Then, in \secref{sec:4_1}, we describe the proposed heterogeneous self-supervised learning and demonstrate its compatibility with existing methods. 
In \secref{sec:3_3}, we demonstrate that the improvements come from the complementarity of heterogeneous architectures. 
\secref{sec:4_2} analyzes what makes a good auxiliary head and discovers that a greater model discrepancy brings more benefits. 
Inspired by this discovery, we propose a quick search strategy to choose the most suitable auxiliary head for a specific base model in \secref{sec:searching} and several simple but effective methods that enlarge the model discrepancy to bring more improvements in \secref{sec:Enlarging_the_Model_Discrepancy}.

\subsection{Preliminaries}
\label{sec:Preliminaries}

The \ourMthd~can be implemented in different forms, 
\eg instance discrimination and masked image modeling. 
In this paper, 
we mainly use the instance discrimination framework as the illustrative example. 
We first briefly recall the common framework of 
instance discrimination. 
Given an image $x$, 
different views of $x$, \ie $x_1$ and $x_2$, 
are generated by different data augmentations. 
Their representations, \ie $z_1$ and $z_2$, 
are extracted by 
teacher and student networks, respectively. 
Then, instance discrimination maximizes 
the similarity between $z_1$ and $z_2$. 
Specifically, the loss function has different forms \cite{caron2021emerging, zhou2021ibot, He_2020_CVPR}, 
and we abstract the loss as ${\mathcal{L}}(z_1^{}, z_2^{})$.

\subsection{Heterogeneous Supervision}
\label{sec:4_1}
Denoting the backbone used by existing methods \cite{caron2021emerging, zhou2021ibot} as 
the base model, 
\ourMthd~utilizes an auxiliary head, 
whose architecture differs from the base model, 
to endow the base model with its missing characteristics. 
The overall pipeline is visualized in \figref{fig:vis_framework}. 
For simplification, 
we refer to the base model/auxiliary head at the teacher and student branches as $f_1$/$h_1$ and $f_2$/$h_2$, respectively. 
Given $x_1$ and $x_2$, 
the base models extract representations $z_1^{b}=f_1^{}(x_1^{})$ and 
$z_2^{b}=f_2^{}(x_2^{})$. 
Then, the auxiliary head takes these representations as input and 
output 
$z_1^h=h_1^{}(z_1^{b})$ and $z_2^h=h_2^{}(z_2^{b})$. 
Since heterogeneous architectures extract $z_1^h$/$z_2^h$ and $z_1^{b}$/$z_2^{b}$, 
the $z_1^h$/$z_2^h$ contains a part of the characteristics that are missing from the $z_1^{b}$/$z_2^{b}$. 
The base model can learn those missing characteristics 
with the loss function $\mathcal{L}(z_1^{h}, z_2^{b})$, 
which pulls $z_1^h$ and $z_2^{b}$ together. 

Meanwhile, 
to guarantee that the auxiliary head can learn meaningful characteristics, 
we also pull representations extracted by auxiliary heads in teacher and student together, 
\ie 
using the loss function $\mathcal{L}(z_1^{h}, z_2^{h})$. 
The base model and the auxiliary head are pre-trained simultaneously, 
and the total loss function $\mathcal{L}$ can be defined as follows:
\begin{equation}\label{eq:L_total}
\mathcal{L} = \mathcal{L}(z_1^{h}, z_2^{b}) + \mathcal{L}(z_1^{h}, z_2^{h}).    
\end{equation}
During pre-training, 
the auxiliary head is serially connected at the end of the base model, 
enabling the former to learn meaningful characteristics with only a few layers. 
Thus, the increased training time and memory costs are negligible. 
After pre-training, 
we remove the auxiliary head 
and only reserve the base model. 

\myPara{Incorporating \ourMthd~into different SSL methods.} 
The proposed \ourMthd~is compatible 
with different self-supervised learning~(SSL) methods, including \MoCo, 
\DINO, 
\iBOT, and \MAE, 
as shown in \tabref{tab:fully_finetune_knn_linear}.
When combined with different methods, 
the loss function defined in \eqref{eq:L_total} 
takes on distinct forms. 
For clustering based methods \cite{caron2021emerging, zhou2021ibot}, 
the representations are transformed into 
probability distributions over $K$ dimensions 
through some projection heads and a softmax function, 
and 
the loss function is defined as follows:
\begin{equation}\label{eq:L_total_cluster}
  \mathcal{L} = -\sum_{i=1}^{K}(z_1^h)_i \log((z_2^{b})_i) - \sum_{i=1}^{K}(z_1^h)_i \log((z_2^{h})_i),   
\end{equation}
where the projection heads and the softmax function are hidden for simplification. 
Additionally, other forms of loss functions can also be combined with \ourMthd, 
\eg InfoNCE \cite{oord2018representation} in contrastive learning \cite{caron2021emerging} and 
reconstruction loss in masked image modeling \cite{he2021masked}. 
For more details, 
please refer to Section 6 of the supplementary material.

\myPara{Analysis for various architectures.} 
\reversion{To validate the effectiveness of the proposed 
\ourMthd, 
we evaluate the impact of different auxiliary heads on the base model. 
In this analysis, 
we aim to explore the effect of diverse architectures on the \ourMthd.
Thus, 
we choose \ResNet, \PoolFormer, \ResMLP, \ConvNext, \ViT, and \Swin~as the auxiliary heads 
due to their diverse architectures. 
For example, 
\ResNet~is a classic convolutional network based on local convolutions, 
and \ConvNext~further adopts large kernel convolutions. 
\ViT~is a transformer network based on global self-attention, 
and \Swin~integrates local attention in the transformer architecture. 
Moreover, 
\PoolFormer~and \ResMLP~adopt 
different modeling mechanisms beyond convolutional and transformer architectures, 
\ie pooling and spatial MLP.}  
As shown in \tabref{tab:differentarch}, 
using the auxiliary head can consistently enhance the base model 
across all pairs\footnote{
For all experiments in 
\secref{sec:method} and \secref{sec:ablation}, 
we adopt the ImageNet-S$_{300}$ dataset \cite{gao2022luss}, 
which contains 300 categories from ImageNet-1K \cite{russakovsky2015imagenet}, 
to save computational costs.
}. 
\bwy{Furthermore, we observe that 
an auxiliary head that is heterogeneous to the base model 
brings more gains than a homogeneous one. 
For example, 
when using ViT as the base model, 
the auxiliary head of the ViT only improves by 0.5\% in Top-1 accuracy. 
In comparison, 
the auxiliary head of the ConvNext brings a 4.2\% improvement in Top-1 accuracy.
These results and observations prove the necessity of the proposed HSSL method.}
\reversion{\bwy{
We also investigate whether relatively weaker auxiliary heads 
can enhance stronger base models. 
\tabref{tab:differentarch_weakhead} shows positive results. 
For example, 
the weaker \PoolFormer~improves the Top-1 accuracy of the \Swin~base model by 0.9\%.
This indicates that our HSSL method is robust to different model architectures and can brings consistent improvements under different settings.}}

\begin{table}[t]
  \centering
  \tablestyle{3.2mm}{1}
  \caption{\reversion{Effects of various auxiliary heads on different base models.}}    
  \begin{tabular}{llcccc} \toprule
    & & \multicolumn{4}{c}{{Base Model}} \\
    & & \tCols{ViT} & \tCols{ResNet} \\ \cmidrule{3-4} \cmidrule{5-6}
    & & Top-1 & Top-5 & Top-1 & Top-5 \\ \hline
    & Baseline    & 67.5 & 84.4 & 63.2 & 84.3 \\ \hline
    \multirow{6}*{\rotatebox{90}{Auxiliary Head}} 
    & \ViT        & 68.0 & 84.7 & 64.0 & 84.3 \\
    & \Swin       & 69.4 & 85.9 & 63.9 & 84.4 \\
    & \PoolFormer & 70.1 & 86.3 & 63.9 & 84.5 \\
    & \ResNet     & 71.7 & 86.9 & 63.5 & 84.3 \\
    & \ResMLP     & 72.6 & {87.8} & {64.4} & {84.9} \\
    & \ConvNext   & {72.7} & 87.6 & 63.7 & 84.4 \\ \bottomrule
  \end{tabular}
  \label{tab:differentarch}
\end{table}

\begin{table}[t]
    \centering
    \tablestyle{6.0mm}{1}
    \caption{\reversion{Weak auxiliary heads also enhance strong base models. 
    Experiments without a declared auxiliary 
    head mean the baselines of corresponding base models.}}    
    \begin{tabular}{llcccc} \toprule
      Base model & Auxiliary head & Top-1 \\
      \midrule
      \ViT & - & 67.5 \\
      \ViT & \ResNet & 71.7 \\
      \ViT & \ResMLP & 72.6 \\
      \midrule
      \ResMLP & - & 58.0 \\
      \ResMLP & \ViT & 59.6 \\
      \midrule
      \Swin & - & 72.8 \\
      \Swin & \PoolFormer & 73.7 \\
      \Swin & \ResMLP & 73.4 \\
      \bottomrule
    \end{tabular}
    \label{tab:differentarch_weakhead}
  \end{table}

\begin{figure*}[t]
    \centering
    \vspace{10pt}
    \begin{overpic}[width=0.98\linewidth]{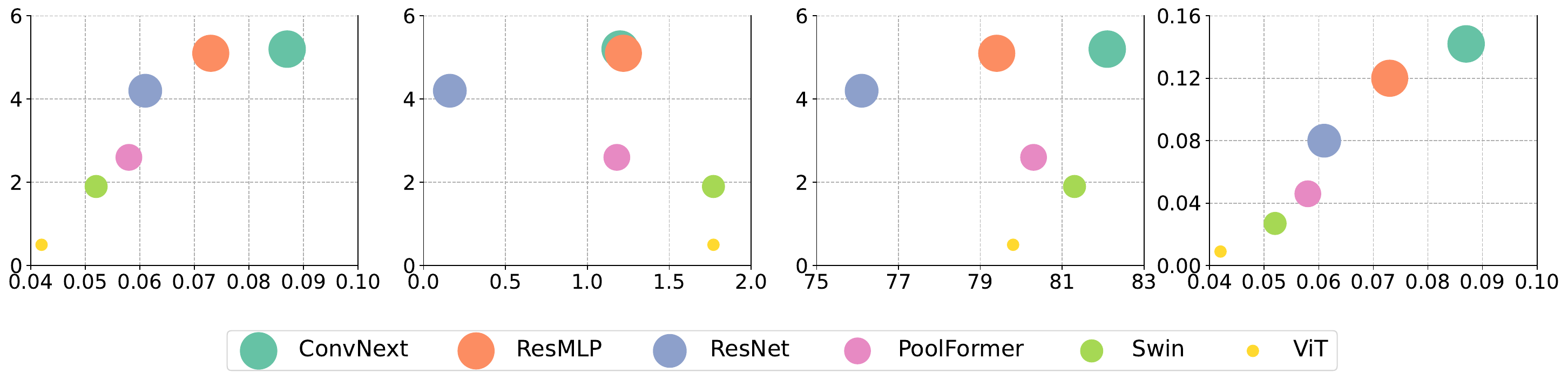}
        \put(-1.0, 07.0){\rotatebox{90}{improvement (Top-1)}}
        \put(24.0, 07.0){\rotatebox{90}{improvement (Top-1)}}
        \put(49.0, 07.0){\rotatebox{90}{improvement (Top-1)}}
        \put(6.5, 4.0){(a) discrepancy}
        \put(54.8, 4.0){(c) capacity (Top-1)}
        \put(26.0, 4.0){(b) number of parameters (M)}
        \put(80.0, 4.0){(d) discrepancy (train)}
        \put(99.0, 7.5){\rotatebox{90}{discrepancy (search)}}
    \end{overpic}
    \caption{
        In (a)-(c), we visualize the relationship between the improvements in the base model~(ViT-S/16) and 
        three factors, 
        including (a) the representation discrepancy between the base model and the auxiliary head, 
        (b) the number of parameters of a 1-layer auxiliary head, 
        (c) The capacity of the architecture that is used to build the auxiliary head. 
        For the capacity of each architecture, 
        we use the supervised classification accuracy on ImageNet-1K, 
        reported in the official paper of each architecture, 
        as a reference to its capacity. 
        In (d), 
        we show a consistent trend 
        between the discrepancies obtained by searching and examining each auxiliary head individually.
        In all figures, the size of the dot is positively related to the improvement 
        brought by the corresponding auxiliary head.
    }
    \label{fig:analysis_arch}
\end{figure*}

\subsection{Heterogeneity Brings Gains} 
\label{sec:3_3}

While the \ourMthd~takes effect across different pairs of the base model and the auxiliary head, 
we further explore how the auxiliary head enhances the base model. 
Specifically, 
we observe that 
the auxiliary head can solve a part of samples that the base model cannot.
To illustrate this, 
we first define sets $B_1$, $B_2$, and $H$, 
which contain the samples that can be 
correctly solved by 
the base model pre-trained by baseline (\DINO), 
the base model pre-trained by \ourMthd, 
and the auxiliary head pre-trained by \ourMthd, 
respectively. 
Meanwhile, 
$U$ means the set that contains all samples of a dataset. 
Then $H \cap (U-B_1)$ contains the samples that 
the auxiliary head can solve but are beyond the capacity of the base model pre-trained by baseline. 
The number of these samples is defined as follows:
\begin{equation}
    N_s = |H \cap (U-B_1)|.
\end{equation} 
Taking ViT as the base model, 
we show that 
the auxiliary head can solve some samples that are beyond the ability of the base model 
in \tabref{tab:unsolved_sample}. 
More importantly, 
an auxiliary head, which can solve more samples unsolved by the base model, 
brings more significant improvements to the base model.

We further investigate whether 
the base model can address those samples in $H \cap (U-B_1)$ under the guide of the auxiliary head.
After pre-training by \ourMthd, 
both the base model and the auxiliary head 
can address some samples that are beyond the capacity of the baseline. 
These samples can be represented 
as $B_2 \cap (U-B_1)$ and $H \cap (U-B_1)$ 
for the base model and the auxiliary head, respectively.  
We notice that there exists a substantial overlap between these two subsets. 
The degree of overlap can be quantified as follows:
\begin{equation}
    {\rm sIoU} = \frac{|B_2 \cap (U-B_1) \cap H \cap (U-B_1)|}{|B_2 \cap (U-B_1)|}.
\end{equation}
\tabref{tab:unsolved_sample} shows the sIoU obtained by 
different auxiliary heads when using ViT as the base model. 
For example, 
there is a 70\% overlap when using \ConvNext~as the auxiliary head. 
The high overlap demonstrates that 
the improvements in the base model can mainly be attributed to 
complementarity and heterogeneity.

\begin{table}[t]
    \centering
    \caption{\reversion{Auxiliary head solves samples that the base model~(ViT) cannot solve. 
    The sIoU and $N_s$ represent the degree of overlap between HSSL and the original version on the correct samples and the number of newly added correct samples from HSSL. Their details are provided in \secref{sec:3_3}.}}    
    \setlength{\tabcolsep}{5.0mm}    
    \renewcommand\arraystretch{1.01}
    \begin{tabular}{lccccccccccccc}
        \toprule
        Auxiliary Head & Top-1 & $N_s$ & sIoU \\
        \midrule
        {\ViT} & 68.0 & 792 & 59.5 \\
        {\Swin} & 69.4 & 854 & 60.7 \\
        {\PoolFormer} & 70.1 & 904 & 60.8 \\
        \ResNet & 71.7 & 1061 & 67.8 \\
        {\ResMLP} & 72.6 & 1270 & {72.9} \\
        {\ConvNext} & {72.7} & {1278} & 70.2 \\
        \bottomrule
        \end{tabular}
    \label{tab:unsolved_sample}
\end{table}

\begin{figure*}[t]
    \centering
    \begin{minipage}[t]{0.33\textwidth}
        \centering
        \begin{overpic}[width=0.90\columnwidth]{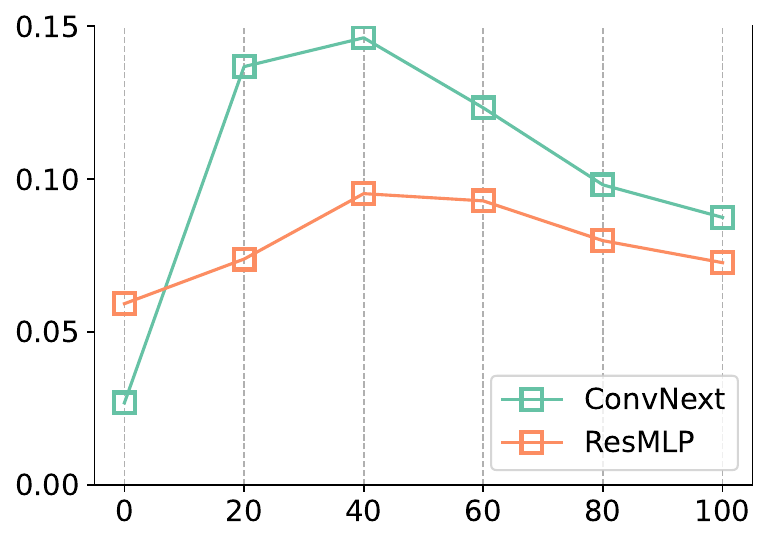}
            \put(47, -3){Epoch}
            \put(-5, 22.5){\rotatebox{90}{Discrepency}}
        \end{overpic}
    \end{minipage}
    \begin{minipage}[t]{0.33\textwidth}
        \centering
        \begin{overpic}[width=0.90\columnwidth]{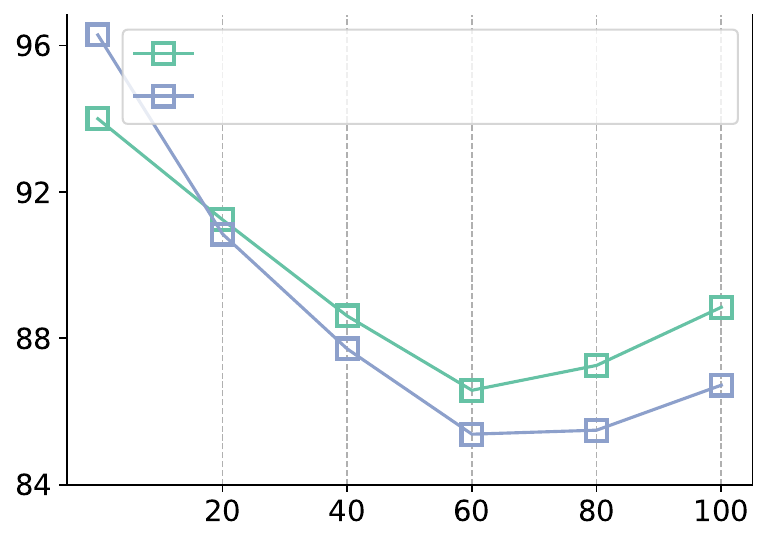}
            \put(47, -3){Epoch}
            \put(-5, 16.5){\rotatebox{90}{CKA similarity}}
            \put(27, 62.5){\footnotesize w. heterogeneous supervision}
            \put(27, 56.8){\footnotesize w/o. heterogeneous supervision}
        \end{overpic}
    \end{minipage}
    \begin{minipage}[t]{0.33\textwidth}
        \centering
        \begin{overpic}[width=0.90\columnwidth]{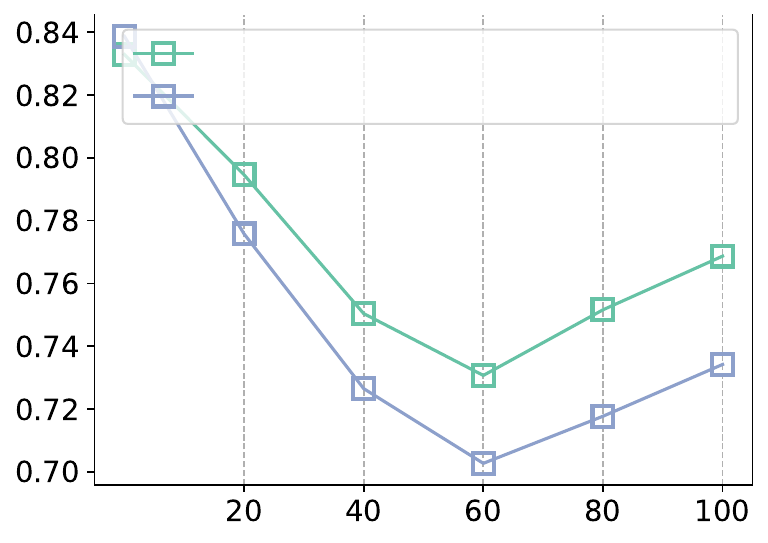}
            \put(47, -3){Epoch}
            \put(-5, 12.5){\rotatebox{90}{Procrustes similarity}}
            \put(27, 62.5){\footnotesize w. heterogeneous supervision}
            \put(27, 56.8){\footnotesize w/o. heterogeneous supervision}
        \end{overpic}
    \end{minipage}
    \caption{
        \reversion{Training dynamics of the discrepancy or similarity 
        between the base model and the auxiliary head during pre-training. 
        Left: The discrepancy $\mathcal{D}$~(defined in \eqref{eq:kl}) 
            between the base model and the auxiliary head 
            when using \ConvNext~or 
            \ResMLP~as the auxiliary head and 
            using \ViT~as the base model. 
        Middle: The feature-level CKA similarity 
            between the base model and the auxiliary head. 
        Right: The feature-level Procrustes similarity 
            between the base model and the auxiliary head.} 
    }
    \label{fig:dynamic}
\end{figure*}

\subsection{Analysis of Model Discrepancy}
\label{sec:4_2}
Different auxiliary heads produce different effects 
for a specific base model, 
as shown in \tabref{tab:differentarch}. 
For ViT, a transformer-based base model, 
using ConvNext as the auxiliary head is more suitable than the others. 
When ResNet is the base model, 
utilizing ResMLP and ViT as auxiliary heads 
can complement global modeling ability and bring more significant improvements. 
The above observation motivates us 
to delve deep into what makes a good auxiliary head. 
By investigating different architectures, 
we discover that 
a more significant discrepancy between the base model and 
the auxiliary head 
brings more gains to the base model. 
This phenomenon inspires us to propose 
a search strategy 
to quickly determine the most suitable auxiliary head 
for a specific base model 
in \secref{sec:searching} and 
several simple but effective methods to magnify the discrepancy 
in \secref{sec:Enlarging_the_Model_Discrepancy}.

\myPara{Model discrepancy.} 
During heterogeneous self-supervised learning, 
the auxiliary head learns a part of characteristics that are 
missing from the base model itself. 
That is to say, there exists a representation discrepancy between the base model and 
the auxiliary head, 
\ie the discrepancy between $z_1^{b}$ and $z_1^h$. 
Taking the self-clustering based methods \cite{caron2021emerging, zhou2021ibot} as an example, 
the $z_1^b$ defined in \secref{sec:Preliminaries} 
means probability distributions over $K$ dimensions. 
Then, we use the Kullback-Leibler divergence to 
measure the discrepancy as follows: 
\begin{equation}
    \mathcal{D} = - (z_1^b)^{T} \log(\frac{z_1^h}{z_1^{b}}), 
    \label{eq:kl}
\end{equation}
where $z_1^{b}$ and $z_1^h$ are extracted 
from the teacher network after pre-training.

\myPara{More significant discrepancy leads to greater improvements.}
Taking ViT-S/16 as an example of the base model, 
in \figref{fig:analysis_arch} (a), 
we show its improvement when it learns from each auxiliary head 
and its discrepancy with each auxiliary head. 
It can be observed that 
there is a positive relationship 
between improvements and discrepancies. 
A more significant discrepancy means the auxiliary head learns
more characteristics that are missing from the base model, 
thus prompting the base model to complement more characteristics.

To further confirm whether the improvement comes from the heterogeneity, 
we analyze other factors, 
including the number of parameters of the auxiliary head and 
the capacity of the architecture used to build the auxiliary head, 
where we use 
the supervised classification accuracy on ImageNet-1K \cite{russakovsky2015imagenet}, 
which is reported by the official paper of each architecture, 
as a reference to the architecture capacity.
As shown in \figref{fig:analysis_arch} (c) and (d), 
both factors have no positive correlation with 
the improvement. 
For example, 
\ViT~has a larger capacity than \ResNet, 
but ResNet is more suitable than ViT when serving 
as the auxiliary head. 
These results 
demonstrate that a greater improvement is not from
a stronger auxiliary head but the heterogeneity.

\myPara{The dynamic of model discrepancy.}
\reversion{Based on the discrepancy analysis, 
we investigate how auxiliary heads influence the base model
during pre-training.
\figref{fig:dynamic} provides detailed insights
into the interaction between the base model and auxiliary heads, 
showing the discrepancy $\mathcal{D}$ (as defined in \eqref{eq:kl}), 
CKA similarity, and Procrustes similarity
between the base model and the auxiliary head, respectively.
From \figref{fig:dynamic}, we observe that the discrepancy initially increases and then decreases during training.
Notably, heterogeneous supervision
significantly amplifies the discrepancy and reduces the similarity, 
as evident in the middle and right panels. 
This observation suggests that heterogeneous supervision
encourages the base model to learn from the auxiliary head.
Moreover, the left panel demonstrates that
using ConvNext as the auxiliary head induces
a larger discrepancy than ResMLP
when ViT is employed as the base model.
This aligns with previous analysis, 
which indicates that a larger discrepancy can 
lead to greater performance improvements.
\bwy{Naturally, the model discrepancy provides us with the possibility to select the optimal auxiliary head for a specific base model.
}}

\subsection{Searching for Suitable Auxiliary Heads} 
\label{sec:searching}

A suitable auxiliary head 
provides more characteristics missing from a specific base model, 
thus complementing the base model better and producing higher improvements. 
However, 
in the unsupervised setting, 
there is no annotated data to evaluate each auxiliary head. 
Inspired by the positive relationship between the discrepancies and improvements, 
we use the model discrepancy to 
determine the most suitable auxiliary head for a specific base model 
via a label-free approach. 
However, 
due to the vast number of candidate auxiliary heads, 
testing candidates one by one is time-consuming. 
Thus, 
we propose an efficient search strategy to 
find the auxiliary head with 
the largest discrepancy to the base model through one quick training.

\begin{figure}[t]
	\centering
	\begin{overpic}[width=0.98\linewidth]{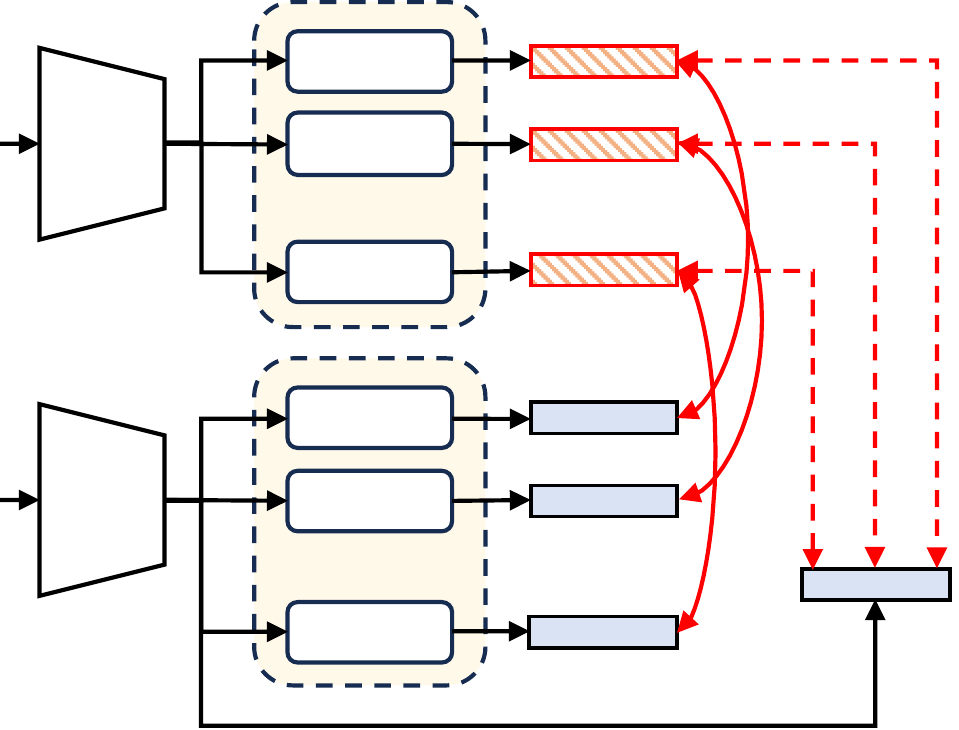}
        \put(37.0, 68.5){$h_1^{1}$}
        \put(60.5, 73.0){$z_1^{h1}$}
        \put(37.0, 60.0){$h_1^{2}$}
        \put(60.5, 64.0){$z_1^{h2}$}
        \put(37.0, 46.5){$h_1^{N}$}
        \put(60.5, 51.0){$z_1^{hN}$}
        \put(38.5, 52.5){\rotatebox{90}{...}}

        \put(37.0, 31.2){$h_2^{1}$}
        \put(60.5, 35.7){$z_2^{h1}$}
        \put(37.0, 22.5){$h_2^{2}$}
        \put(60.5, 27.0){$z_2^{h2}$}
        \put(37.0, 9){$h_2^{N}$}
        \put(60.5, 13.2){$z_2^{hN}$}
        \put(38.5, 15.5){\rotatebox{90}{...}}

        \put(93.0, 9.5){$z_2^{b}$}

        \put(9.0, 23.3){$f_2$}
        \put(9.0, 60.3){$f_1$}

        \put(-0.5, 27.0){$x_2$}
        \put(-0.5, 64.0){$x_1$}
	\end{overpic}
	\caption{
        \reversion{
        Illustration of the quick search strategy. 
        Given $N$ distinct architectures, 
        we construct $N$ different auxiliary heads, 
        where $h_{1/2}^i$ represents the auxiliary head 
        built using $i$-th architecture. 
        The subscripts 1 and 2 indicate teacher and student branches, respectively. 
        In the figure, 
        the \red{red} dotted lines and solid lines 
        correspond to the loss of the first and second terms of \eqref{eq:L_total_search_each}. 
        Projection heads are omitted from for clarity.}}
    \label{fig:search_pipeline}
\end{figure}

\begin{table}[t]
    \begin{minipage}[t]{1.0\linewidth}
        \centering
        \setlength{\tabcolsep}{4.0mm}
        \caption{\reversion{Cooperation of multiple auxiliary heads when using ViT as the base model. `$\mathcal{D}$' represents the discrepancy degree between the auxiliary head and the base model.}}
        \begin{tabular}{lccccccc}
            \toprule
            {Auxiliary Head} & $\mathcal{D}$ & Top-1 & Top-5 \\
            \midrule
            ResMLP & 7.3e-2 & 72.6 & 87.8 \\
            ConvNext & 8.7e-2 & 72.7 & 87.6 \\
            ConvNext+ResMLP & 11.0e-2 & {73.7} & {88.2} \\
            \bottomrule
        \end{tabular}
        \label{tab:multiple_head}
    \end{minipage}
\end{table}

\myPara{Quick Search Strategy.}
\reversion{
Unlike the standard \ourMthd~architecture, 
which employs a single auxiliary head, 
we arrange all the candidate auxiliary heads in parallel during training, 
as shown in \figref{fig:search_pipeline}.
This allows each auxiliary head to independently perform 
heterogeneous self-supervised learning without interference.
Suppose there are 
$N$ candidate auxiliary heads, 
each corresponding to a distinct architecture.
For the inputs $x_1$ and $x_2$, 
we first send them to the 
base models in teacher and student branches 
to generate representations $z_1^{b}=f_1(x_1)$ and 
$z_2^{b}=f_2(x_2)$, respectively.
Then, 
in the teacher branch, 
the $N$ auxiliary heads further 
process $z_1^{b}$ and 
produce heterogeneous representations 
$\{z_{1}^{hi} \mid i \in [0, N - 1] \}$. 
Similarly, the student branch generates 
$\{z_{2}^{hi} \mid i \in [0, N - 1] \}$.
For the $i$-th auxiliary head, 
we define the loss function 
like \eqref{eq:L_total} as follows:
\begin{equation}
    \mathcal{L}_{}^{hi} = \mathcal{L}(z_1^{hi}, z_2^{b}) + \mathcal{L}(z_1^{hi}, z_2^{hi}).
    \label{eq:L_total_search_each}
\end{equation}
The overall loss function across all auxiliary heads is given by:
\begin{equation}
    \mathcal{L}_{}^{s} = \frac{1}{N}\sum_{i=0}^{N-1}  \mathcal{L}_{}^{hi}.
    \label{eq:L_total_search_all}
\end{equation}
In practice, 
the features in \eqref{eq:L_total_search_each} 
are passed through independent projection heads 
before calculating the loss, 
as commonly adopted in prior work \cite{caron2021emerging, he2021masked}. 
During searching, 
we apply an independent projection head for each auxiliary head. 
To minimize mutual interference, 
the representations $z_{1/2}^b$ of the base model 
are passed into separate projection heads when 
paired with different auxiliary heads. 
For clarity, these projection heads are omitted in \eqref{eq:L_total_search_each} and \eqref{eq:L_total_search_all}.}

\reversion{
After training with \eqref{eq:L_total_search_all}, 
we calculate the discrepancy between the base model and 
each auxiliary head. 
For the $i$-th head, 
the discrepancy $\mathcal{D}_i$ is computed as:
\begin{equation}
\mathcal{D}_i = - (z_1^b)^{T} \log(\frac{z_1^{hi}}{z_1^{b}}).
\label{eq:L_total_search_model_d}
\end{equation}
Finally, the auxiliary head with the largest discrepancy is selected:
\begin{equation}
    \arg \max_{i} \mathcal{D}_i, 
    \label{eq:L_total_search_argmax}
\end{equation} 
where the $i$-th auxiliary head is identified as 
the most complementary auxiliary head to the base model.
\bwy{Therefore, the optimal auxiliary head for the base model can be rapidly identified.}
}

\myPara{Searching time.}
Compared to examining each auxiliary head through multiple training, 
the proposed search strategy requires only one training. 
Because we use a very shallow auxiliary head, 
the base model accounts for most of the computational budget 
during training. 
As a result, 
when there are six auxiliary heads, 
training with all of them simultaneously, 
\ie the proposed search strategy, 
requires only 1.4$\times$ training time than 
training with one. 
Thus, 
the search strategy requires only about $\frac{1.4 \times 1}{1 \times 6} \approx 23\%$ 
of the time required by examining 
all auxiliary heads one by one. 
Meanwhile, 
we empirically discover that using only 10\% of the training data is 
enough for searching, 
further reducing the search time significantly.

\myPara{Searching results.}
Taking ViT as the base model, 
we analyze the relative relationship
of its discrepancies with different auxiliary heads. 
As shown in \figref{fig:analysis_arch} (b), 
the relative relationship obtained during searching 
aligns with that obtained by testing each auxiliary head individually, 
verifying the effectiveness of the search strategy. 

\begin{table}[t]
    \begin{minipage}[t]{1.0\linewidth}
        \centering
        \setlength{\tabcolsep}{4.0mm}
        \caption{\reversion{Analysis of the shortcut connection in the auxiliary head. 
        Here, we adopt ViT as the base model and ConvNeXt as the auxiliary head.}}
        \begin{tabular}{lccccccc}
            \toprule
            {The first shortcut} & $\mathcal{D}$ & Top-1 & Top-5 \\
            \midrule
            Preservation & 5.6e-2 & 71.0 & 86.8 \\
            Removal & 8.7e-2 & {72.7} & {87.6} \\
            \bottomrule
        \end{tabular}
        \label{tab:depth_lnn}
    \end{minipage}
\end{table}

\begin{figure}[t]
    \centering
    \subfigure{
        \begin{overpic}[width=0.465\columnwidth]{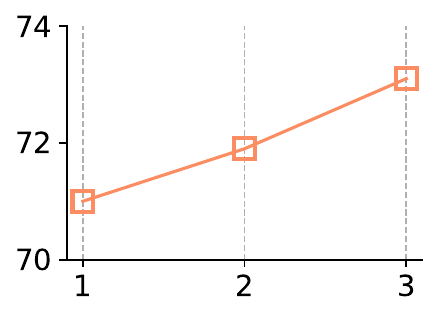}
        \put(45, -3){Depth}
        \put(-4, 30.5){\rotatebox{90}{Top-1}}
        \end{overpic}
    }
    \subfigure{
        \begin{overpic}[width=0.465\columnwidth]{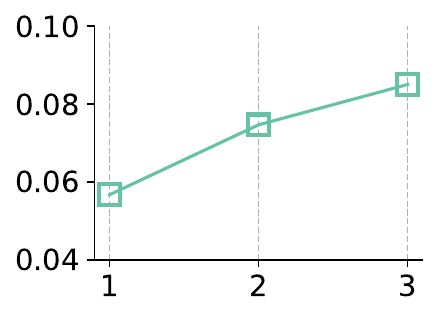}
        \put(49, -3){Depth}
        \put(-5, 37.0){\rotatebox{90}{$\mathcal{D}$}}
        \end{overpic}
    }
    \caption{
    Influence of network depth in the auxiliary head. 
    We take the ViT as the base model and ConvNext as the auxiliary head. 
    }
    \label{fig:difference_depth_kernel}
\end{figure}

\begin{figure*}[t]
	\centering
	\begin{overpic}[width=1.0\linewidth]{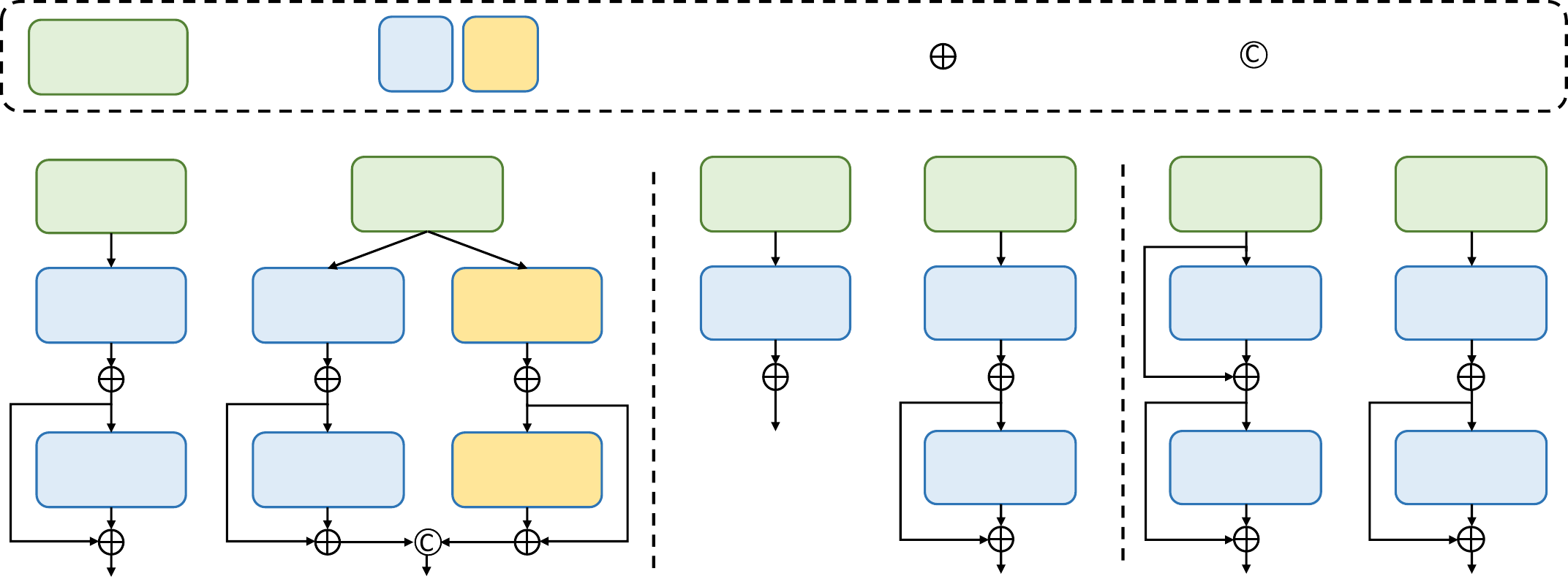}
        \put(13.0, 33.1){\small Base model}
        \put(35.0, 34.3){\small Auxiliary heads with}
        \put(35.0, 32.3){\small different architectures}
        \put(62.0, 33.3){\small Sum}
        \put(82.0, 33.3){\small Concatenation}
        \put(5, -1.0){\small (a) Cooperation of multiple auxiliary heads}
        \put(44.5, -1.0){\small (b) Deepening the auxiliary head}
        \put(75, -1.0){\small (c) Removing the ﬁrst shortcut}
	\end{overpic}
	\caption{
        Three strategies to enlarge the model discrepancy. 
        For each strategy, the left shows the baseline and 
        the right shows the specific strategy.}
    \label{fig:enlarge_d}
\end{figure*}

\subsection{Enlarging the Model Discrepancy} 
\label{sec:Enlarging_the_Model_Discrepancy}

\secref{sec:4_2} demonstrates that a larger model discrepancy 
can bring more improvement gains.
Inspired by this observation, 
we propose three simple but effective technologies to magnify such discrepancy and further boost the performance.

\myPara{Cooperation of multiple auxiliary heads.}
\reversion{The base model only learns limited 
characteristics from a specific auxiliary head. 
Inspired by the principles of ensemble learning, 
where combining multiple models can bring performance improvements, 
we try to combine multiple auxiliary heads built by 
distinct architectures 
to complement more characteristics that are missing from the base model.} 
Specifically, 
supposing there are $n$ auxiliary heads composed of different architectures, 
we represent them as $\{h_1^i | i \in [1, n]\}$ and $\{h_2^i | i \in [1, n]\}$ 
in teacher and student, respectively. 
With the representations $z_{1/2}^b$ produced by the base model, 
each auxiliary head $h_{1/2}^i$ processes them and generates 
representations $h_{1/2}^i(z_{1/2}^b)$. 
\reversion{As shown in \figref{fig:enlarge_d} (a), these representations are combined 
as follows:}
\begin{equation}
    z_{1/2}^{hc} = {\rm concat}(\{h_{1/2}^i(z_{1/2}^b) | i \in [1, n]\}), 
\end{equation}
where ${\rm concat}$ means the concatenation along the channel dimension. 
Then, 
we substitute these representations into \eqref{eq:L_total} and 
get the new loss functions as follows:
\begin{equation}
    \mathcal{L} = \mathcal{L}(z_1^{hc}, z_2^{b}) + \mathcal{L}(z_1^{hc}, z_2^{hc}).
    \label{eq:L_total_multi}
\end{equation}
Compared to a single auxiliary head, 
multiple ones can 
provide more characteristics required by the base model. 
As shown in \tabref{tab:multiple_head}, 
when using two auxiliary heads simultaneously, \ie \ConvNext~and \ResMLP, 
we achieve greater improvements 
than just using ConvNext or ResMLP. 
\bwy{This demonstrates that our HSSL method can achieve further improvements through cooperation of multiple auxiliary heads.}

\myPara{Deepening the auxiliary head.} 
\reversion{Several works \cite{licascade, Xie_2023_CVPR} have observed that 
representations learned across different layers of a model 
exhibit discrepancies, 
with larger gaps between layers resulting in greater disparities. 
Meanwhile, 
a deeper network 
can learn more powerful representations \cite{he2016deep}.
Thus, 
we are inspired to deepen the auxiliary head 
to make it learn more characteristics different from 
the base model. 
This is implemented by simply stacking more blocks on the auxiliary head, 
as shown in \figref{fig:enlarge_d} (b).
The results in \figref{fig:difference_depth_kernel} 
verify that deepening the auxiliary head from one to three layers 
enlarges the discrepancy and brings greater improvements.}

\myPara{Removing the first shortcut connection.}
In most architectures, 
the shortcut connection is utilized by default 
to ensure convergence. 
However, in our \ourMthd, 
we observe that the shortcut in the first layer of the auxiliary head 
reduces the discrepancy between the base model and the auxiliary head. 
To illustrate this argument, we take a two-layer auxiliary head as an example, 
where the two layers are represented as 
$F_1$ and $F_2$, respectively. 
We use $z$ to represent the output of the base model. 
When remaining the first shortcut, 
the auxiliary head outputs $z+F_1(z)+F_2(z+F_1(z))$. 
In comparison, 
the auxiliary head outputs $F_1(z)+F_2(F_1(z))$ 
when we remove the first shortcut. 
We can observe that the former directly adds $z$, 
\ie the output of the base model, 
to the output of the auxiliary head, 
thus reducing the model discrepancy. 
This phenomenon is further illustrated in the supplementary material.
Thus, 
we remove the first shortcut connection, 
as shown in \figref{fig:enlarge_d} (c). 
This approach enlarges the discrepancy and leads to more significant improvements, 
as shown in \tabref{tab:depth_lnn}.

\section{Experiments}
\label{sec:exps}

\begin{table}[t]
    \centering
    \caption{
    \reversion{Cooperating the proposed \ourMthd~with 
    various architectures and frameworks. 
    We report Top-1 on the validation set of ImageNet-1K \cite{russakovsky2015imagenet}, 
    using the evaluation protocols of 
    fully fine-tuning, linear probing, and $k$-NN, respectively. 
    $^\dag$ means that we use the multi-crop strategy \cite{caron2021emerging} 
    with 2 global crops of $224 \times 224$ and 10 local crops of $96\times 96$. 
    }}    
    \setlength{\tabcolsep}{1.0mm}
    \begin{tabular}{lccccc}
        \toprule
        & {Architecture} & Epoch\tablefootnote{\label{foot:effevtive_epoch}For a fair comparison, we report effective epochs \cite{zhou2021ibot} that account for actual images used during pre-training. 
        For iBOT and DINO that use 10 local crops, the number of effective epochs is four times the number of actual epochs.}
        & Fine-tuning & Linear & $k$-NN \\
        \midrule
        \MAE & \tRows{ViT-S/16} & \tRows{400} & 80.4 & - & - \\
        \MAE+\ourMthd & & & {80.8} & - & - \\
        \midrule
        \MoCo & \tRows{ViT-S/16} & \tRows{200} & - & 65.3 & 57.4 \\
        \MoCo+\ourMthd & & & - & {65.7} & {58.1} \\
        \midrule
        \DINO & \tRows{ViT-S/16} & \tRows{200} & - & 67.9 & 61.2 \\
        \DINO+\ourMthd & & & - & {70.8} & {65.5} \\
        \midrule
        \iBOT & \tRows{ViT-S/16} & \tRows{200} & - & 71.3 & 65.2 \\
        \iBOT+\ourMthd & & & - & {72.6} & {67.3} \\
        \midrule
        \DINO$^\dag$ & \tRows{ViT-S/16} & \tRows{400} & - & 74.6 & 70.9 \\
        \DINO$^\dag$+\ourMthd & & & - & {75.7} & {72.5} \\
        \midrule
        \iBOT$^\dag$ & \tRows{ViT-S/16} & \tRows{400} & 80.9 & 74.4 & 71.5 \\
        \iBOT$^\dag$+\ourMthd & & & {81.3} & {76.5} & {72.8} \\
        \midrule
        \AttMask$^\dag$ & \tRows{ViT-S/16} & \tRows{400} & - & 76.1 & 72.8 \\
        \AttMask$^\dag$+\ourMthd & & & - & {76.7} & {73.1} \\
        \midrule
        \iBOT$^\dag$ & \tRows{ViT-B/16} & \tRows{400} & 83.3 & 77.8 & 74.0 \\
        \iBOT$^\dag$+\ourMthd & & & {83.8} & {79.4} & {75.3} \\
        \midrule
        \iBOT$^\dag$ & \tRows{ViT-B/16} & {1600} & 84.0 & 79.5 & 77.1 \\
        \iBOT$^\dag$+\ourMthd & & {600} & {84.1} & {79.6} & 76.0 \\
        \midrule
        \reversion{\MFF} & \tRows{\reversion{ViT-B/16}} & \tRows{\reversion{300}} & \reversion{83.3} & - & - \\
        \reversion{\MFF+\ourMthd} & & & \reversion{83.6} & - & - \\
        \midrule
        \reversion{\DINO} & \tRows{\reversion{Swin-T}} & \tRows{\reversion{200}} & - & \reversion{69.2} & \reversion{60.1} \\
        \reversion{\DINO+\ourMthd} & & & - & \reversion{71.2} & \reversion{65.1} \\
        \midrule
        \reversion{\DINO} & \tRows{\reversion{PVT-Small}} & \tRows{\reversion{200}} & - & \reversion{67.7} & \reversion{61.3} \\
        \reversion{\DINO+\ourMthd} & & & - & \reversion{69.6} & \reversion{64.4} \\
        \bottomrule
        \end{tabular}
    \label{tab:fully_finetune_knn_linear}
\end{table}

\begin{table}[t]
    \centering
    \caption{\reversion{Comparison with previous methods using ViT-B/16 \cite{dosovitskiy2020vit}. 
    $^\dag$ means the usage of a pre-trained perceptual codebook for the tokenization.}
    }    
    \setlength{\tabcolsep}{4mm}
    \begin{tabular}{lccc}
        \toprule
        & Epochs$^{\ref{foot:effevtive_epoch}}$ & Fine-tuning & Linear \\
        \midrule
        \MoCo & 600 & 83.2 & 76.7 \\
        \DINO & 1600 & 83.6 & 78.2 \\
        SimMIM \cite{Xie_2022_CVPR} & 800 & 83.8 & 56.7 \\
        \MAE & 1600 & 83.6 & 68.0 \\
        \iBOT & 400 & 83.3 & 77.8 \\
        MaskFeat \cite{Wei_2022_CVPR} & 1600 & 84.0 & - \\
        BootMAE \cite{dong2022bootstrapped} & 800 & 84.2 & 66.1 \\
        SdAE \cite{chen2022sdae} &  300 & 84.1 & 64.9 \\
        BEiT \cite{bao2021beit} & 800 & 83.2 & 56.7 \\
        SiameseIM \cite{tao2023siamese} & 400 & 83.7 & 76.8 \\
        MOKD \cite{song2023multimode} & 400 & - & 78.0 \\
        LocalMIM \cite{Wang_2023_CVPR} & 1600 & 84.0 & - \\
        MFF \cite{Liu_2023_ICCV} & 800 & 83.6 & - \\
        CIM \cite{fang2023corrupted} & 300 & 83.3 & -  \\
        ConMIM \cite{yi2022masked} & 800 & 83.7 & 39.3 \\
        ccMIM \cite{zhang2023contextual} & 300 & 83.6 & 66.9 \\
        ccMIM \cite{zhang2023contextual} & 800 & 84.2 & 68.9 \\
        PeCo \cite{dong2023peco}$^\dag$ &300 & 84.1 & -  \\
        SERE \cite{sere} & 400 & 83.7 & 77.9 \\
        \midrule
        \iBOT+\ourMthd & 400 & 83.8 & 79.4 \\
        \iBOT+\ourMthd & 600 & 84.1 & 79.6 \\
        \MFF+\ourMthd & 300 & 83.6 & - \\
        \bottomrule
        \end{tabular}
    \label{tab:comparison}
  \end{table}

\begin{table}[t]
    \centering
    \caption{\reversion{Comparison on semantic segmentation.
    We fine-tune UperNet \cite{Xiao_2018_ECCV} with the ViT-B/16 \cite{dosovitskiy2020vit} as the backbone on the ADE20K \cite{Zhou_2017_CVPR} dataset, 
    following existing works \cite{he2021masked, zhou2021ibot}.}}
    \setlength{\tabcolsep}{4mm}
    \begin{tabular}{lccc}
        \toprule
        & Architecture & Epochs$^{\ref{foot:effevtive_epoch}}$ & mIoU \\
        \midrule
        \MoCo & \multirow{14}{*}{ViT-B/16} & 600 & 47.2 \\
        \DINO & & 1600 & 46.8 \\
        \MAE & & 1600 & 48.1 \\
        BootMAE \cite{dong2022bootstrapped} & & 800 & 49.1 \\
        SdAE \cite{chen2022sdae} & & 300 & 48.6 \\
        BEiT \cite{bao2021beit}$^\ddag$ & & 800 & 45.6 \\
        SiameseIM \cite{tao2023siamese} & & 400 & 49.6 \\
        MixedAE \cite{Chen_2023_CVPR} & & 800 & 48.7 \\
        LocalMIM \cite{Wang_2023_CVPR} & & 1600 & 49.5 \\
        MFF \cite{Liu_2023_ICCV} & & 800 & 48.6 \\
        ConMIM \cite{yi2022masked} & & 800 & 46.0 \\
        ccMIM \cite{zhang2023contextual} & & 800 & 47.7 \\
        PeCo \cite{dong2023peco}$^\dag$ & & 300 & 48.5 \\
        SERE \cite{sere} & & 800 & 50.0 \\
        \midrule
        \iBOT & \thRows{ViT-B/16} & 400 & 47.9 \\
        \iBOT & & 1600 & 50.0 \\
        \iBOT+\ourMthd & & 400 & 50.3 \\
        \midrule
        \iBOT & \thRows{ViT-S/16} & {400} & 45.2 \\
        \iBOT & & {3200} & 45.4 \\
        \iBOT+\ourMthd & & {400} & 46.1 \\
        \bottomrule
        \end{tabular}
    \label{tab:semantic_segmentation}
  \end{table}

  \begin{table}[t]
    \centering
    \caption{\reversion{Comparison on object detection and instance segmentation 
    with ViT-B/16. 
    We fine-tune the models on the COCO \cite{lin2015microsoft} dataset 
    and report AP$^{\rm m}$ as segmentation mask AP and 
    AP$^{\rm b}$ as bounding box AP, respectively.}}
    \setlength{\tabcolsep}{3mm}
    \begin{tabular}{lcccc}
        \toprule
        & {Architecture} & Epochs$^{\ref{foot:effevtive_epoch}}$ & AP$^{\rm m}$ & AP$^{\rm b}$ \\
        \midrule
        \DINO & \thRows{ViT-B/16} & 1600 & 43.4 & 50.1 \\
        \MAE & & 1600 & 44.3 & 51.3 \\
        SERE \cite{sere} & & 400 & 43.8 & 50.7 \\
        \midrule
        \MFF & \multirow{2}*{ViT-B/16} & 300 & 43.2 & 50.0 \\
        \MFF+\ourMthd & & 300 & 43.7 & 50.5 \\
        \midrule
        \iBOT & \multirow{2}*{ViT-B/16} & 400 & 43.2 & 50.1 \\
        \iBOT+\ourMthd & & 400 & 44.0 & 51.0 \\
        \midrule
        \iBOT & \multirow{2}*{ViT-B/16} & 1600 & 44.2 & 51.2 \\
        \iBOT+\ourMthd & & 600 & 44.3 & 51.4 \\
        \bottomrule
        \end{tabular}
    \label{tab:det}
  \end{table}

\begin{table}[t]

    \begin{minipage}[t]{1.0\linewidth}
        \centering
        \caption{\reversion{Cross-domain transferring learning on RAW object detection. 
        The models are fine-tuned on AODRaw \cite{li2024aodraw}, which collects RAW images for object detection. 
        Apart from the Average Precision~(AP) \cite{lin2015microsoft}, 
        we also report AP$_{\rm 75}$ and AP$_{\rm 50}$ 
        at the IoU threshold of 0.75 and 0.50. 
        AP$_{\rm s}$, AP$_{\rm m}$, and AP$_{\rm l}$ mean the AP for small, medium, and large objects.}}    
        \setlength{\tabcolsep}{1.8mm}
        \begin{tabular}{lcccccccc}
            \toprule
            & Architecture & Epochs$^{\ref{foot:effevtive_epoch}}$ & AP & AP$_{\rm 50}$ & AP$_{\rm 75}$ \\
            \midrule
            \DINO & \multirow{2}*{Swin-T \cite{liu2021Swin}} & 200 & 28.9 & 45.7 & 30.2 \\
            \DINO+\ourMthd & & 200 & 29.5 & 46.5 & 30.8 \\
            \bottomrule
        \end{tabular}
        \label{tab:aodraw}
    \end{minipage}
    \begin{minipage}[t]{1.0\linewidth}
        \centering
        \caption{\reversion{Transfer learning on more image classification benchmarks, including CIFAR \cite{Krizhevsky_2009_17719} and iNaturalist \cite{Horn_2018_CVPR}.}}    
        \setlength{\tabcolsep}{0.8mm}
        \begin{tabular}{lcccccccc}
            \toprule
            & Architecture & Epochs$^{\ref{foot:effevtive_epoch}}$ & Cifar$_{100}$ & INat$_{18}$ & INat$_{19}$ \\
            \midrule
            \iBOT & \thRows{ViT-B/16} & 400 & 92.1 & 74.0 & 78.4 \\
            \iBOT & & 1600 & 92.2 & 74.6 & 79.6 \\
            \iBOT+\ourMthd & & {400} & {92.2} & {75.2} & {79.7} \\
            \bottomrule
            \end{tabular}
        \label{tab:transfer_cls}
    \end{minipage}
\end{table}

\subsection{Experimental Settings}
\label{sec:settings}
\reversion{We integrate \ourMthd~with a wide range of self-supervised methods, 
including MoCov3 \cite{Chen_2021_ICCV}, \DINO, \AttMask, 
\iBOT, \MAE~and \MFF.} 
For each method, 
we follow its official implementation. 
During pre-training, 
we adopt ViT-S/16 or ViT-B/16 architecture as the base model, 
and the auxiliary head uses the
ConvNext architecture unless otherwise specified. 
In the auxiliary head, 
we default the depth to 3 and 
remove the shortcut connection at the first layer. 
More details about pre-training and fine-tuning are shown in the supplementary material.

\subsection{Experimental Results}
\label{sec:performance}

\myPara{Image classification on ImageNet-1K.}
We first fully fine-tune the base models on ImageNet-1K 
and compare the classification performance, 
as shown in \tabref{tab:fully_finetune_knn_linear}. 
Using ViT-B/16, 
\ourMthd~improves by 0.5\% in Top-1 accuracy 
over \iBOT~when pre-training for 400 epochs. 
With 600 epochs, 
\ourMthd~can achieve 84.1\% Top-1 accuracy, 
even outperforming iBOT of 1600 epochs. 
\reversion{We also combine \ourMthd~with masked image modeling~(MIM) based methods, 
and the implementation details are shown in the supplementary material. 
In \tabref{tab:fully_finetune_knn_linear}, 
\ourMthd~enhances MAE by 0.4\% on Top-1 accuracy 
after pre-training ViT-S/16 for 400 epochs. 
Compared to \MFF, 
we advance the performance by 0.3\% Top-1 accuracy 
after pre-training ViT-B/16 for 300 epochs.}

We also evaluate the effectiveness of \ourMthd~using 
$k$-NN and linear probing on the ImageNet-1K dataset. 
\reversion{As shown in \tabref{tab:fully_finetune_knn_linear}, 
\ourMthd~consistently improves various methods, 
including instance discrimination based~(\eg \DINO~and \MoCo), 
and hybrid methods that combine instance discrimination 
with MIM~(\eg \iBOT~and \AttMask).} 
For example, 
when pre-training ViT-B/16 by 400 epochs, 
\ourMthd~advances iBOT
by 1.6\% in linear probing accuracy. 
Meanwhile, 
\ourMthd~can achieve comparative performances over iBOT 
with even fewer epochs~(600 vs. 1600 epochs). 
These results show that 
\ourMthd~enhances the ability of classification and 
is orthogonal to existing representation learning methods. 
\reversion{
Moreover, 
\tabref{tab:fully_finetune_knn_linear} highlights that \ourMthd~can enhance 
different transformer architectures, 
extending beyond the plain vision transformer \cite{dosovitskiy2020vit}. 
For example, 
\ourMthd~improves the Swin-T \cite{liu2021Swin} and PVT-Small \cite{wang2021pyramid} 
by 2.0\% and 1.9\% in linear probing accuracy, respectively, 
after pre-training for 200 epochs. 
}

\reversion{Finally, 
we directly compare our \ourMthd~with prior methods, 
as shown in \tabref{tab:comparison}. 
Compared to instance discrimination based methods, 
our method pre-trained for 600 epochs 
achieves 84.1\% and 79.6\% Top-1 accuracy 
on fully fine-tuning and linear probing, respectively, 
outperforming methods such as \iBOT~that requires 1600 epochs to achieve 79.5\% linear probing accuracy. 
Even with a shorter pre-training schedule of 400 epochs, 
\ourMthd~still surpasses iBOT by 0.5\% and 1.6\% 
on fully fine-tuning and linear probing, respectively.
Although some MIM-based methods, 
\eg ccMIM \cite{zhang2023contextual} and BootMAE \cite{dong2022bootstrapped}
deliver competitive performance in fine-tuning, 
they lag in linear probing accuracy and 
may exhibit limited effectiveness on downstream tasks, 
as shown in \tabref{tab:semantic_segmentation}. 
In contrast, \ourMthd~achieves superior performance across 
both linear probing, fine-tuning, and downstream tasks.}

\myPara{Transfer learning on image classification.}
Besides ImageNet-1K, 
we also transfer the pre-trained base models to 
other classification datasets, 
including CIFAR \cite{Krizhevsky_2009_17719} and iNaturalist \cite{Horn_2018_CVPR}.
As shown in \tabref{tab:transfer_cls}, 
\ourMthd~brings consistent improvements across different datasets, 
demonstrating superior transferability.

\myPara{Transfer learning on semantic segmentation.}
We use UperNet \cite{Xiao_2018_ECCV} as the segmentation model 
for evaluating semantic segmentation performance. 
\reversion{Following prior works \cite{zhou2021ibot}, 
we fine-tune the models on the ADE20K \cite{Zhou_2017_CVPR} dataset. 
As shown in \tabref{tab:semantic_segmentation}, 
\ourMthd~achieves 50.3\% mIoU 
after pre-training ViT-B/16 for 400 epochs. 
Notably, 
\ourMthd~outperforms \iBOT, 
which requires 1600 epochs of pre-training to 
achieve similar results.
Using ViT-S/16 pre-trained for 400 epochs, 
\ourMthd~also 
advances \iBOT~by 0.9\% mIoU.} 
These results highlight the effectiveness of \ourMthd~on 
dense prediction.

\begin{table}[t]
    \centering
    \caption{Semi-supervised classification on ImageNet-1K \cite{russakovsky2015imagenet}. 
    We utilize linear and $k$-NN classifiers with 1\%/10\% labels and report the Top-1 accuracy. }    
    \setlength{\tabcolsep}{2.0mm}
    \begin{tabular}{lcccccc}
        \toprule
        & {Architecture} & {Epochs$^{\ref{foot:effevtive_epoch}}$} & 1\% &  10\% \\
        \midrule
        \iBOT & \tRows{ViT-B/16} & 400 & 64.8 & 76.3 \\
        \iBOT+\ourMthd & & 400 & {66.1} & {76.8} \\
        \bottomrule
    \end{tabular}
    \label{tab:semi}
\end{table}

\begin{table}[t]
    \centering
    \caption{Semi-supervised semantic segmentation on ImageNet-S \cite{gao2022luss}. 
    We report the mIoU on the val and test sets. 
    The PT means pre-trained weights initiate the model, 
    and FT means fully fine-tuned weights initiate the model, respectively.}    
    \setlength{\tabcolsep}{0.9mm}
    \begin{tabular}{lcccccc}
        \toprule
        & \tRows{Architecture} & \tRows{Epochs$^{\ref{foot:effevtive_epoch}}$} & \tCols{ImageNet-S$_{\rm PT}$} & \tCols{ImageNet-S$_{\rm FT}$} \\
        \cmidrule(lr){4-5} \cmidrule(lr){6-7} 
        & & & val & test & val & test \\
        \midrule
        \iBOT & \thRows{ViT-B/16} & 400 & 48.3 & 47.8 & 62.6 & 63.0 \\
        \iBOT & & 1600 & 50.5 & 50.1 & - & - \\
        \iBOT+\ourMthd & & {400} & {51.5} & {51.1} & {63.5} & {63.1} \\
        \bottomrule
    \end{tabular}
    \label{tab:semiseg}
\end{table}

\begin{table}[t]
    \begin{minipage}[t]{1.0\linewidth}
        \centering
        \caption{Unsupervised semantic segmentation on ImageNet-S \cite{gao2022luss}. 
        919/300/50 mean the ImageNet-S/ImageNet-S$_{300}$/ImageNet-S$_{50}$ datasets, respectively. 
        We follow the pipeline and setting proposed in \cite{gao2022luss} 
        and report mIoU on the val and test sets. 
        Here, we do not adopt the multi-crop strategy for the representation learning.}    
        \setlength{\tabcolsep}{1.5mm}
        \begin{tabular}{lcccccc}
            \toprule
            & Datasets & {Architecture} & Epochs$^{\ref{foot:effevtive_epoch}}$ & val & test \\
            \midrule
            \iBOT & \tRows{50} & \tRows{ViT-S/16} & \tRows{400} & 46.2 & 45.1 \\
            \iBOT+\ourMthd & & & & {54.4} & {54.5} \\
            \midrule
            \iBOT & \tRows{300} & \tRows{ViT-S/16} & \tRows{200} & 22.2 & 22.4 \\
            \iBOT+\ourMthd & & & & {26.6} & {26.0} \\
            \midrule
            \iBOT & \tRows{919} & \tRows{ViT-S/16} & \tRows{200} & 12.2 & 11.3 \\
            \iBOT+\ourMthd & & & & {14.0} & {13.6} \\
            \bottomrule
            \end{tabular}
        \label{tab:luss}
    \end{minipage}
    \begin{minipage}[t]{1.0\linewidth}
        \centering
        \caption{
        Time and memory usage during pre-training on an 8-GPU machine, 
        with a batch size of 256 and 10 multi-crops of 96$\times$96.}    
        \setlength{\tabcolsep}{1.9mm}
        \begin{tabular}{lccccc}
            \toprule
            & {Architecture} & Epochs$^{\ref{foot:effevtive_epoch}}$ & Time (h) & Memory (G)\\
            \midrule
            \iBOT & \tRows{ViT-B/16} & \tRows{400} & 82.7 & 18.3 \\
            \iBOT+\ourMthd & & & 94.5 & 21.4 \\
            \bottomrule
            \end{tabular}
        \label{tab:time}
    \end{minipage}
\end{table}

\myPara{Transfer learning on instance segmentation.}
Following \cite{zhou2021ibot}, 
we use Cascade Mask R-CNN \cite{Cai_2018_CVPR} 
to implement instance segmentation and object detection. 
\reversion{As shown in \tabref{tab:det}, 
\ourMthd~advances \iBOT~by 0.8\% ${\rm AP^m}$ and 0.9\% ${\rm AP^b}$ 
with just 400 epochs of pre-training. 
Compared to \MFF, 
\ourMthd~delivers a 0.5\% improvement in 
both ${\rm AP^m}$ and ${\rm AP^b}$. 
Notably, \ourMthd~also reduces training costs, 
achieving superior performance by 
lowering the required pre-training epochs from 1600 (as in \iBOT) to just 600.} 
Furthermore, \tabref{tab:aodraw} shows that our method significantly improves object detection accuracy in the RAW domain~\cite{li2024aodraw}, demonstrating strong cross-domain generalization.

\myPara{Cross-domain transferring.}
\reversion{We also evaluate the pre-trained models on the 
AODRaw \cite{li2024aodraw} dataset, 
designed for object detection in the RAW domain. 
The RAW domain presents a significant domain gap 
compared to the sRGB domain on which we pre-train models. 
The results show that our \ourMthd~achieves a 0.6\% improvement in AP 
when pre-training Swin-T for 200 epochs, 
highlighting its strong cross-domain generalization capability.}

\myPara{Semi-supervised learning.} 
Collecting annotations 
requires huge costs. 
Semi-supervised learning 
can reduce the demand for expensive annotations. 
Thus, 
we also evaluate the ability of \ourMthd~in 
semi-supervised classification and 
semantic segmentation. 
We follow the paradigm in \cite{zhou2021ibot} for semi-supervised classification to 
fine-tune the pre-trained base models with a part of labels.
As shown in \tabref{tab:semi}, 
\ourMthd~improves by 1.3\% and 0.5\% in Top-1 accuracy over \iBOT~when using 1\% and 10\% training labels, 
respectively. 
For semi-supervised semantic segmentation, 
we fine-tune the base models on the ImageNet-S \cite{gao2022luss} dataset, 
in which 919 categories and 9190 labeled images are included. 
\tabref{tab:semiseg} reports the mIoU on the val and test sets. 
We can observe that \ourMthd~significantly improves \iBOT~
by 4.7\% and 4.2\% mIoU on the val and test sets. 

\myPara{Unsupervised semantic segmentation.}
We evaluate the pre-trained base models with unsupervised semantic segmentation. 
For training, we follow the pipeline proposed in \cite{gao2022luss} and 
consider three datasets \cite{gao2022luss}, 
\ie ImageNet-S$_{50}$, ImageNet-S$_{300}$, and ImageNet-S datasets. 
As shown in \tabref{tab:luss}, 
\ourMthd{}~outperforms iBOT by 1.8\% mIoU 
on the ImageNet-S dataset. 
The results in semi-supervised and 
unsupervised learning show that 
\ourMthd~benefits the perception and recognition 
in the absence of labels.

\myPara{Time and memory usage.} 
\tabref{tab:time} shows the time and memory usage required by \iBOT~and our \ourMthd. 
Compared to the baseline, 
the \ourMthd~only increases negligible computation costs 
because 
the serial connection between the base model and the auxiliary head 
enables the auxiliary head to extract helpful representations 
with just a 
few layers.

\section{Ablation and Analysis}
\label{sec:ablation}

We perform ablation studies by pre-training models for 100 actual epochs on the ImageNet-S$_{300}$ 
to save computation costs. 
By default, we set the depth of the auxiliary head to 1.
We evaluate the performance 
by reporting {knn classification accuracy (Cls.) on the ImageNet and 
segmentation mIoU (Seg.) on the ImageNet-S.} 

\myPara{Effect of the supervision manner.}
After connecting the auxiliary head, 
we investigate whether to use 
the base model itself or the auxiliary head to 
guide the base model, 
where the former is homogeneous 
and the latter is heterogeneous. 
As shown in \tabref{tab:direction}, 
the heterogeneous manner outperforms the homogeneous manner, achieving 5.5\% higher mIoU and 4.7\% higher Top-1 accuracy.
These results verify 
that heterogeneous supervision is essential, 
allowing the base model to learn complementary characteristics from the auxiliary head.

\begin{table}[t]
    \begin{minipage}[t]{1.0\linewidth}
        \centering
        \setlength{\tabcolsep}{4.5mm}
        \caption{Ablation for the supervision manner on the base model. 
        \tb{B} and \tb{A} mean the base model and the auxiliary head, respectively. 
        \tb{A$\rightarrow$B} means that the auxiliary head supervises the base model.
        \tb{B$\rightarrow$B} means the base model supervises itself.}
        \begin{tabular}{lccccc}
            \toprule
            & {Seg.} & \tCols{Cls.} \\
            \cmidrule(lr){2-2} \cmidrule(lr){3-4}
            & mIoU & Top-1 & Top-5 \\
            \midrule
            A$\rightarrow$A & 16.1 & 26.5 & 48.0 \\
            A$\rightarrow$A + B$\rightarrow$B  & 31.4 & 68.0 & 86.4 \\
            A$\rightarrow$A + \tb{A$\rightarrow$B}  & {36.9} & {72.7} & {87.6} \\
            \bottomrule
        \end{tabular}
        \label{tab:direction}
    \end{minipage}
    \begin{minipage}[t]{1.0\linewidth}
        \centering
        \caption{Ablation for the structure of the auxiliary head.}
        \setlength{\tabcolsep}{4.0mm}
        \begin{tabular}{lcccc}
            \toprule
            & {Seg.} & \tCols{Cls.} \\
            \cmidrule(lr){2-2} \cmidrule(lr){3-4}
            & mIoU & Top-1 & Top-5 \\
            \midrule
            MLP & 35.8 & 70.0 & 86.3 \\
            Token Mixer & 36.3 & 70.1 & 86.4 \\
            MLP + Token Mixer & {36.9} & {72.7} & {87.6} \\
            \bottomrule
        \end{tabular}
        \label{tab:structure}
    \end{minipage}
    \begin{minipage}[t]{1.0\linewidth}
        \centering
        \caption{Ablation for the shared projection and not shared projection.}
        \setlength{\tabcolsep}{5.0mm}
        \begin{tabular}{ccccccc}
            \toprule
            \tRows{Shared proj.} & {Seg.} & \tCols{Cls.} \\
            \cmidrule(lr){2-2} \cmidrule(lr){3-4}
            & mIoU & Top-1 & Top-5 \\
            \midrule
            \ding{52} & 35.8 & 72.3 & 87.5 \\
            \ding{56} & {36.9} & {72.7} & {87.6} \\
            \bottomrule
        \end{tabular}
        \label{tab:shared}
    \end{minipage}
\end{table}

\myPara{Structure of the auxiliary head.}
We use a unified framework for different auxiliary heads, 
which includes a token mixer and an MLP block. 
Here, we take ConvNext as an example and 
evaluate the effect of the token mixer and MLP block. 
The results presented in \tabref{tab:structure} show that 
the token mixer plays a more crucial role, 
leading to improvements of 0.6\% mIoU and 2.6\% Top-1 accuracy compared to the MLP block.

\myPara{Whether to share the projections.}
Before calculating the losses, 
self-supervised learning methods usually 
process the teacher/student representations through some projection heads.
\tabref{tab:shared} investigates whether to share the projections
between the base model and the auxiliary head. 
The results indicate that not sharing the projections 
provides an advantage of 1.1\% mIoU and 0.4\% Top-1 accuracy. 
Due to the different architectures, 
the representations between the base model and the auxiliary head have discrepancies, 
and not sharing the projections allows greater flexibility in processing the discrepancy.

\myPara{Parallel or serial connection for the auxiliary head.}
We can connect the auxiliary head and the base model
in serial or parallel. 
For the parallel connection, 
we use the entire ConvNext-Tiny as the auxiliary head that directly takes the images as input. 
In contrast, 
the serial connection allows the auxiliary head to extract rich information 
with just a few layers.
As shown in \tabref{tab:parallel}, 
the parallel arrangement requires about 2.32 $\times$ training time 
compared to the serial connection. 
Moreover, 
using serial connection achieves better performances than 
parallel arrangement, 
achieving better computational efficiency.

\begin{table}[t]
    \begin{minipage}[t]{1.0\linewidth}
        \centering
        \caption{Ablation for parallel and serial connections of the auxiliary head. 
        We use a depth of 3 for serial connection. 
        We show the multiples relative to the baseline for the time and memory costs.}
        \setlength{\tabcolsep}{2.5mm}
        \begin{tabular}{lcccccc}
            \toprule
            & {Seg.} & \tCols{Cls.} & \tCols{Computation cost} \\
            \cmidrule(lr){2-2} \cmidrule(lr){3-4} \cmidrule(lr){5-6}
            & mIoU & Top-1 & Top-5 & {time} & {memory} \\
            \midrule
            baseline & 29.3 & 67.5 & 84.4 & $\times$1.00 & $\times$1.00 \\
            parallel & 34.6 & 72.8 & 87.2 & $\times$2.53 & $\times$2.25 \\
            serial & {37.1} & {73.9} & {88.4} & $\times$1.09 & $\times$1.12\\
            \bottomrule
        \end{tabular}
        \label{tab:parallel}
    \end{minipage}
    \begin{minipage}[t]{1.0\linewidth}
        \centering
        \setlength{\tabcolsep}{3.0mm}   
        \vspace{10pt}
        \caption{Utilizing heterogeneous self-supervision on different 
        granularity when using ViT, taking \iBOT~as an example.}
        \begin{tabular}{cccccc}
            \toprule
            \tRows{Image-level} & \tRows{Patch-level} & {Seg.} & \tCols{Cls.} \\
            \cmidrule(lr){3-3} \cmidrule(lr){4-5}
            & & mIoU & Top-1 & Top-5 \\
            \midrule
            \ding{56} & \ding{56} & 42.3 & 75.1 & 89.3 \\
            \ding{52} & \ding{56} & 46.2 & 75.8 & 89.4 \\
            \ding{52} & \ding{52} & {46.7} & {76.0} & {89.5} \\
            \bottomrule
        \end{tabular}
        \label{tab:ablation_clstoken_patch}
    \end{minipage}
    \begin{minipage}[t]{1.0\linewidth}
        \centering
        \vspace{10pt}
        \caption{Comparison with the strategy of deep-to-shallow~(DTS) \cite{gao2022luss}.}
        \setlength{\tabcolsep}{5.0mm}
        \begin{tabular}{lcccc}
            \toprule
            & {Seg.} & \tCols{Cls.} \\
            \cmidrule(lr){2-2} \cmidrule(lr){3-4}
            & mIoU & Top-1 & Top-5 \\
            \midrule
            baseline & 29.3 & 67.5 & 84.4 \\
            +DTS & 30.5 & 68.6 & 85.4 \\
            +\ourMthd{} & {36.9} & {72.7} & {87.6} \\
            \bottomrule
        \end{tabular}
        \label{tab:d2s}
    \end{minipage}
\end{table}

\myPara{Cls token and patch token.}
Some methods \cite{zhou2021ibot, attmask, sere} 
calculate losses on 
different granularity simultaneously. 
Taking \iBOT, 
which considers losses on 
both image-level and patch-level, 
as an example, 
\tabref{tab:ablation_clstoken_patch} 
shows the effects when using \ourMthd~on different granularity. 
Note that 
when only using \ourMthd~on the image-level, 
the pixel-level self-supervision 
is only used between the base models of teacher and student. 
It can be seen that 
the base model can learn the majority of the helpful information 
from the auxiliary 
via only image-level supervision. 
Meanwhile, 
learning with pixel-level supervision 
also brings further improvement. 
These results show that 
we can save computational costs 
by only applying \ourMthd~on the image-level.

\myPara{Comparison with deep-to-shallow.}
The deep-to-shallow enhances the representations of a shallow layer with supervision from a deeper layer
within a homogeneous architecture. 
As shown in \tabref{tab:d2s}, 
this strategy only leads to a slight improvement in the ViT, 
likely because 
the deep and shallow layers in ViT are highly similar \cite{raghu2021do}, 
making the supervision lack diversity.
In contrast, 
the heterogeneous self-supervised learning prompts the ViT to learn diverse knowledge, 
achieving significant improvements of 
6.4\% mIoU and 4.1\% Top-1 accuracy over the DTS.

\section{Conclusion}
In this paper, we propose heterogeneous self-supervised learning~(\ourMthd{}). 
Specifically, we enforce a base model to learn 
from an auxiliary head whose architecture is heterogeneous to the base model, 
endowing the base model with some characteristics that are missing from itself. 
Furthermore, 
we discover that 
the discrepancy between the base model and the auxiliary head is 
positively correlated to the improvements brought by \ourMthd. 
This positive correlation 
motivates us to propose an efficient search strategy 
that finds the most 
suitable auxiliary head for a specific model 
and several simple but effective designs to enlarge the model discrepancy. 
We show that \ourMthd{} is orthogonal 
to different self-supervised learning methods and 
boosts the performance 
on various downstream tasks, 
including image classification, semantic segmentation, object detection, and instance segmentation. 

\myPara{Limitations and further works.}
\reversion{\ourMthd~has been successfully integrated with various self-supervised learning methods 
to enhance performance. 
However, certain methods, 
such as those in\cite{gao2022towards}, 
which utilize representations extracted from a frozen pre-trained model 
as learning targets, are less straightforward to adapt. 
Future work could focus on developing more general or 
specialized approaches to effectively incorporate heterogeneous 
representation learning with a broader range of methods.
Additionally, 
\ourMthd~measures model discrepancy using the 
Kullback-Leibler divergence between the probability distributions output by different models. 
While this metric is well-suited for clustering-based self-supervised learning, 
alternative metrics, such as CKA similarity\cite{kornblith2019similarity}, 
could be explored to evaluate discrepancy for other self-supervised learning methods that output representations.
Regarding the searching strategy, 
future research can also aim to design more accurate and 
efficient search strategies that accommodate a wider range of models, 
including those with complex structures, 
further broadening the applicability and scalability.}

\bibliographystyle{IEEEtran}
\bibliography{egbib}

\vfill

\end{document}